\documentclass[letterpaper]{article} % DO NOT CHANGE THIS
\usepackage{AnonymousSubmission/LaTeX/aaai2026}  % DO NOT CHANGE THIS

\usepackage{times}  % DO NOT CHANGE THIS
\usepackage{helvet}  % DO NOT CHANGE THIS
\usepackage{courier}  % DO NOT CHANGE THIS
\usepackage[hyphens]{url}  % DO NOT CHANGE THIS
\usepackage{graphicx} % DO NOT CHANGE THIS
\usepackage{natbib}  % DO NOT CHANGE THIS AND DO NOT ADD ANY OPTIONS TO IT
\usepackage{caption} % DO NOT CHANGE THIS AND DO NOT ADD ANY OPTIONS TO IT
\usepackage{multirow}
\usepackage{booktabs}      
\usepackage{subcaption}

\usepackage{algorithm}
\usepackage{algorithmic}

\usepackage{newfloat}
\usepackage{listings}
\DeclareCaptionStyle{ruled}{labelfont=normalfont,labelsep=colon,strut=off} % DO NOT CHANGE THIS
\floatstyle{ruled}
\newfloat{listing}{tb}{lst}{}
\floatname{listing}{Listing}
\usepackage{bibentry}
\title{Test-Time-Matching: Decouple Personality, Memory, and Linguistic Style in LLM-based Role-Playing Language Agent}

\author{
    Xiaoyu Zhan,
    Xinyu Fu,
    Hao Sun,
    Yuanqi Li,
    Jie Guo,
    and Yanwen Guo\thanks{Corresponding author.\\   Code available: https://github.com/ZhanxyR/TTM}\\
}
\affiliations {
    \textsuperscript{\rm}Nanjing University\\
    \texttt{\{zhanxy, xinyu.fu, hao.sun\}@smail.nju.edu.cn} \\
    \texttt{\{yuanqili, guojie, ywguo\}@nju.edu.cn}
}

\begin{document}

\maketitle

\begin{abstract}
The rapid advancement of large language models (LLMs) has enabled role-playing language agents to demonstrate significant potential in various applications. However, relying solely on prompts and contextual inputs often proves insufficient for achieving deep immersion in specific roles, particularly well-known fictional or public figures. On the other hand, fine-tuning-based approaches face limitations due to the challenges associated with data collection and the computational resources required for training, thereby restricting their broader applicability. To address these issues, we propose Test-Time-Matching (TTM), a training-free role-playing framework through test-time scaling and context engineering. TTM uses LLM agents to automatically decouple a character’s features into personality, memory, and linguistic style. Our framework involves a structured, three-stage generation pipeline that utilizes these features for controlled role-playing. It achieves high-fidelity role-playing performance, also enables seamless combinations across diverse linguistic styles and even variations in personality and memory. We evaluate our framework through human assessment, and the results demonstrate that our method achieves the outstanding performance in generating expressive and stylistically consistent character dialogues.
\end{abstract}

% Uncomment the following to link to your code, datasets, an extended version or similar.
% You must keep this block between (not within) the abstract and the main body of the paper.
% \begin{links}
    % \link{Code}{https://github.com/ZhanxyR/TTM}
    % \link{Datasets}{https://aaai.org/example/datasets}
    % \link{Extended version}{https://aaai.org/example/extended-version}
% \end{links}

\section{Introduction}

Benefiting from the rapid advancement of large language models (LLMs), Role-Playing Language Agents (RPLAs) have become a prominent area of research, spurring a diverse array of commercial applications \cite{CharacterAI, talkie, maoxiang, whisper2025}. Recent research has explored various approaches to enhance the role-playing capabilities of LLMs, including parametric training methods \cite{shao2023character, lu2024large, zhou2023characterglm, li2023stylized, yu2024neeko, yang2024crafting} and non-parametric prompting techniques \cite{tu2023characterchat, li2023chatharuhi, huang2024emotional, li2024stylechat}. 

However, fine-tuning-based approaches are constrained by the difficulties in data collection and the substantial computational resources required for training, limiting their widespread applicability. Meanwhile, prompts and contextual inputs alone are often insufficient for achieving immersive playing in specific roles. In contrast to prior work, this study focuses on decoupling linguistic style from cognitive traits such as personality and memory, aiming to achieve more modular and controllable role-playing behavior in language agents with more detailed context learning.

\begin{figure}[t]
\centering
\includegraphics[width=1\columnwidth]{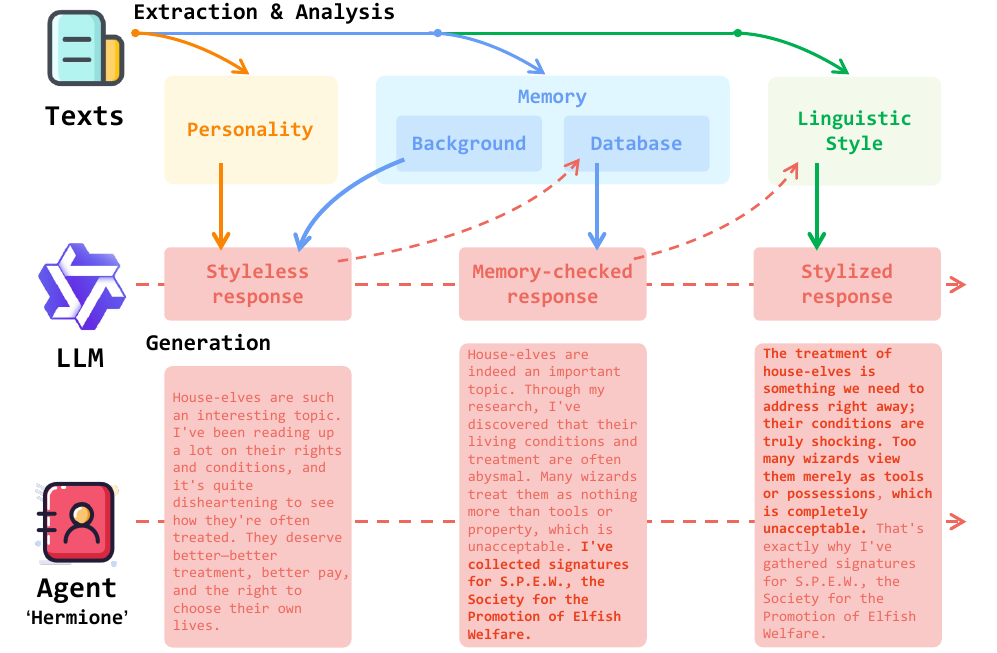} 
\caption{Decouple personality, memory and linguistic style in RPLA. TTM adopts a structured three-stage generation pipeline to generate high-fidelity responses for specific characters. (Hermione is known for her serious, firm, undoubted and slightly preachy tone in the \textit{Harry Potter} series, especially when she promoted the S.P.E.W.)}
\label{fig:teaser}
\end{figure}

In RPLA, linguistic style refers to the distinctive language habits and expressive preferences associated with predefined character profiles. We posit that linguistic style can be conceptually and functionally decoupled from cognitive tendencies in dialogues, which are largely shaped by personality and memory. Cognitive tendencies primarily govern decision, making processes in conversational contexts, such as the selection of affirmative or negative responses, or the retrieval and reference of specific historical facts during interaction. In contrast, linguistic style pertains to the distinct verbal expressions through which characters convey their attitudes or stances. For instance, when aligning with a particular viewpoint, one character might express agreement through a directly utterance such as "Wow, I agree with you." whereas another may employ a more stubborn formulation like "Well, well, well, I have to admit that you're right.". Many studies \cite{li2023chatharuhi, zhou2023characterglm, chen2024persona} have recognized the importance of these three components in RPLAs, however, they often address them in an integrated or holistic pipeline.

In contrast, the decoupling of linguistic style from cognitive tendencies in dialogue modeling offers significant benefits. This approach enables more interpretable and controllable behavioral analysis. It attributes the model outputs to either cognitive decisions or linguistic formulation, thereby supporting more effective optimization. It also facilitates a smoother alignment between the styleless cognitive tendencies in role-playing with the latent demographic patterns \cite{chen2024persona} inherent in pretrained LLMs. Moreover, this decoupling enhances controllability and personalization, enabling seamless combinations across diverse linguistic styles and even variations in personality and memory.

For this purpose, we propose Test-Time Matching (TTM), a novel framework aims to achieve high-fidelity role-playing performance through test-time scaling and context engineering. TTM leverages LLM agents to perform fine-grained extraction of personality, memory, and linguistic style from input texts, enabling a structured and interpretable representation of role-specific features. Central to TTM is its ability to decouple these features during generation, which is achieved through a structured three-stage generation pipeline (in Figure \ref{fig:teaser}). In the first stage, a styleless response is generated based solely on the personality and background. This ensures that the core behavioral and dispositional characteristics of the role are embedded in the output, independent of linguistic style. In the second stage, detailed memory, implemented through the retrieval-augmented generation (RAG), is utilized to correct factual inconsistencies or add contextually relevant details to the response. Finally, the extracted style features are applied to the content-rich but styleless response. Through targeted style transfer, the output is adapted to align with the target role’s distinctive verbal patterns, tone, and expressive traits, thereby improving overall role fidelity.

\begin{figure}[t]
\centering
\includegraphics[width=1\columnwidth]{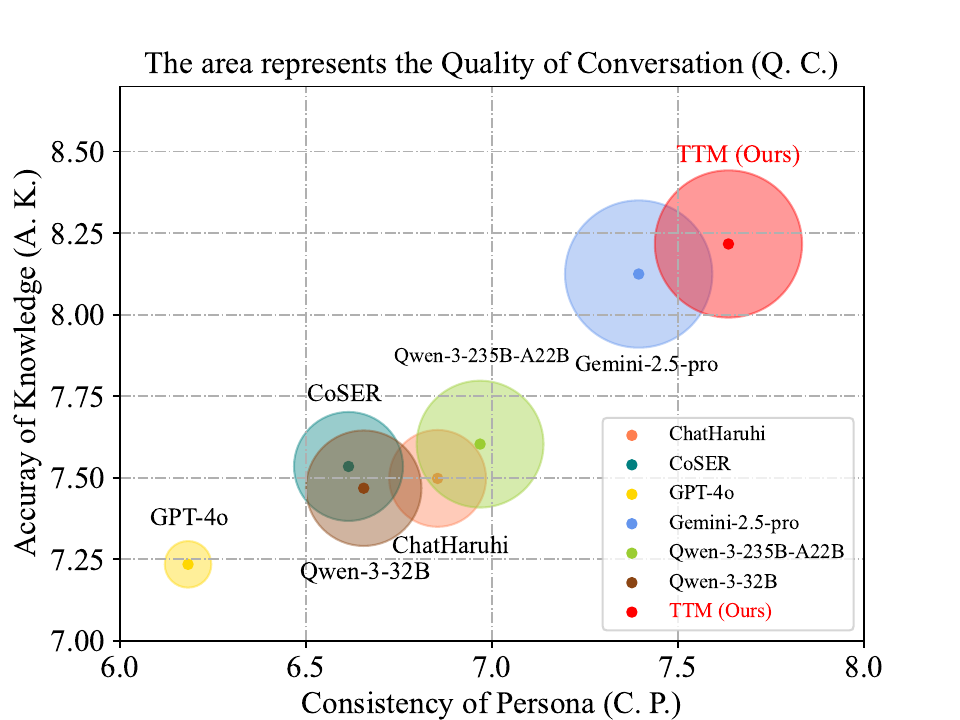} 
\caption{We present the average ratings from the LLM-as-Judge approach, general participants, and linguistics experts (with a maximum score of 10). The proposed TTM, built upon Qwen-3-32B, demonstrates superior overall performance compared to other methods.}
\label{fig:score_visualization}
\end{figure}

In order to evaluate the effectiveness of TTM in enhancing role-playing capabilities, we conduct human assessment in which human raters are asked to express their preferences among conversations generated by different methods. The experimental results indicate that TTM significantly improves the ability of LLMs to play specific roles, not only in terms of role fidelity and role-consistent knowledge, but also in the accurate reproduction of the target linguistic style (in Figure \ref{fig:score_visualization}).

In summary, our contributions are threefold.

\begin{itemize}
    \item We introduce a fully automated framework for extracting structured role-related information from texts, and constructing RPLAs for corresponding characters without parametric training.
    \item We decouple personality, memory, and linguistic style in RPLAs to enable controllable generation and personalized customization.
    \item We employ test-time scaling and context engineering to achieve high-fidelity role-playing, with implementation based on a structured three-stage dialogue generation pipeline.
\end{itemize}

\section{Related Work}

\begin{figure*}[t]
\centering
\includegraphics[width=2\columnwidth]{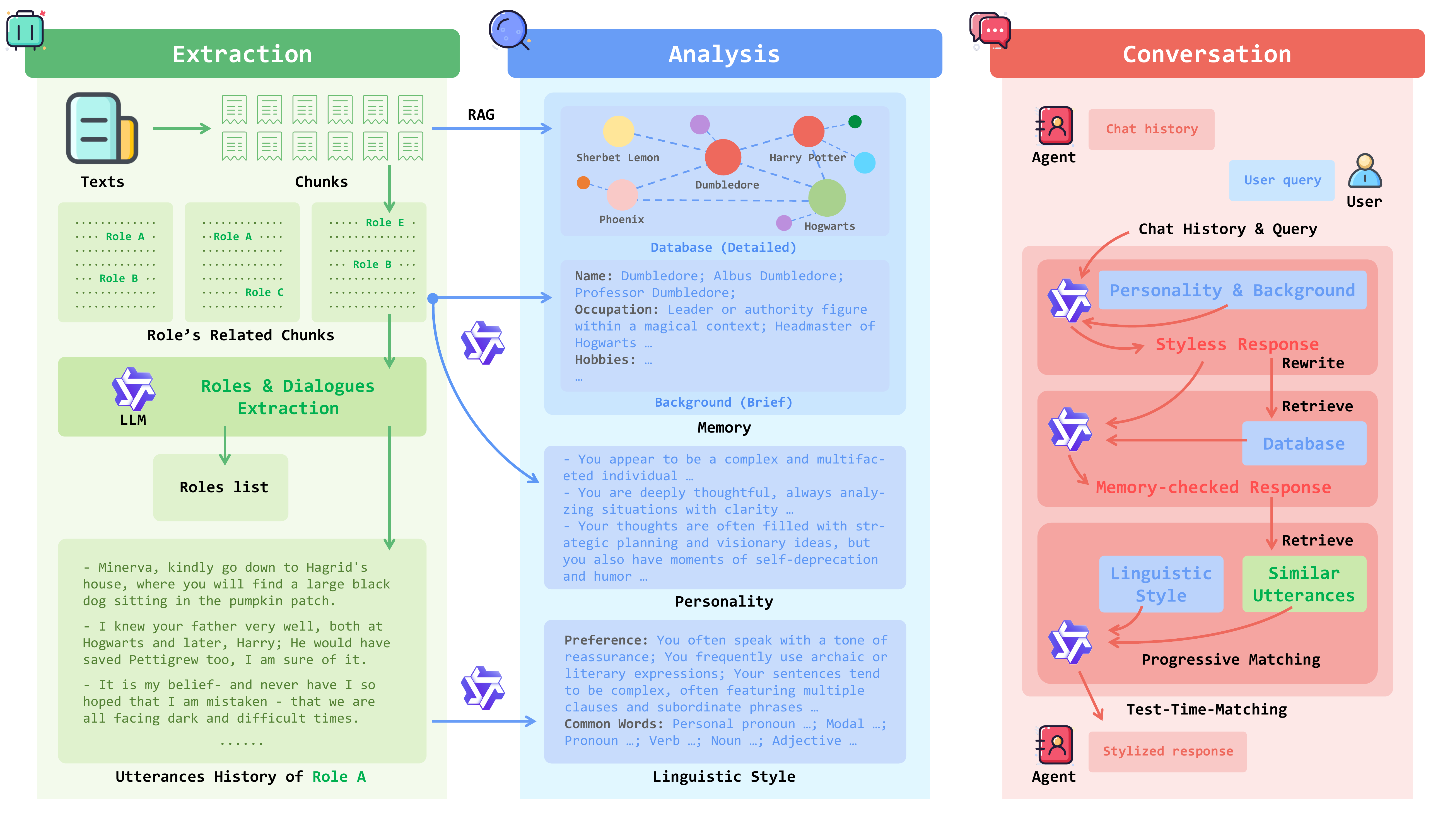} 
\caption{Pipeline overview. TTM first automatically extracts role-relevant information from textual inputs and subsequently analyzes the decoupled features (personality, memory, and linguistic style). During role-playing, TTM employs test-time scaling, which consists of a structured three-stage generation pipeline, to achieve high-fidelity performance.}
\label{fig:pipeline}
\end{figure*}

\subsection{Role-Playing Language Agents}

RPLAs are designed to immersively embodying specific characters. By simulating the background, cognitive patterns, and linguistic habits of the target character, these agents aim to provide users with an immersive and contextually coherent interactive experience. RPLAs hold significant potential across a wide range of applications, including emotional companionship \cite{talkie, CharacterAI}, game non-player characters (NPCs) \cite{buongiorno2024pangea, whisper2025}, scriptwriting \cite{ryu2025cinema, han2024ibsen}, and beyond. In recent years, rapid advancements in LLMs have driven substantial progress in the development and capabilities of RPLAs.

\textbf{Datasets.} To facilitate the development and training of RPLAs, a lot of research has explored various methods for constructing high-quality datasets. These approaches can be broadly categorized into manual creation \cite{zhou2023characterglm}, automatic extraction techniques \cite{li2023chatharuhi, wang2025coser}, and the generation of synthetic data using LLMs based on provided examples or prompts \cite{zhou2023characterglm, wang2024rolellm, lu2024large, li2024stylechat}. 

\textbf{Parametric Training.} Plenty of studies have focused on enhancing the performance and fidelity of RPLAs through parametric training strategies. These approaches typically incorporate structured character profiles as foundational data to guide dialogue generation \cite{zhou2023characterglm, wang2024rolellm, lu2024large, yang2024crafting, wang2025coser}. In addition, many frameworks integrate historical dialogues and experiential memory to preserve conversational continuity \cite{li2023stylized, wang2024rolellm, li2024stylechat, wang2025coser}. Narrative elements such as plots and world-building contexts are also employed to situate interactions within coherent fictional or real-world scenarios \cite{shao2023character, yang2024crafting, wang2025coser}. Among these efforts, \citet{shao2023character} and \citet{zhang2024thinking} introduce mechanisms aimed at constraining model knowledge to mitigate hallucinations and off-character responses.

\textbf{Non-parametric Prompting.} Another line of research investigates the realization of RPLAs through non-parametric prompting techniques, which aim to achieve role-consistent behaviors without additional training. These approaches primarily include prompt engineering, context learning, and retrieval-augmented generation \cite{tu2023characterchat, li2023chatharuhi, wang2024rolellm, huang2024emotional}.

\textbf{Evaluation.} In parallel to advancements in the construction and training of RPLAs, several studies \cite{tu2024charactereval, wang2024incharacter, lu2024large, dai2024mmrole} have focused on developing automatic evaluation frameworks to assess the performance and fidelity of RPLAs in specific roles. Nonetheless, expert human assessment continues to serve as the most authoritative and persuasive measure in the evaluation of role-playing performance.

\subsection{Text-Style Transfer}

Text-style transfer is also a widely researched problem \cite{hu2017toward, lample2019multiple, dai2019style, lee2021enhancing, toshevska2021review, zhu2023storytrans, roy2023conversation, pan2024unsupervised, zhang2024distilling, tao2024cat}. The challenge of text-style transfer lies in the lack of parallel corpora, making it difficult to apply direct supervised learning approaches \cite{dai2019style}. Moreover, the quality of style transfer is inherently hard to quantify and evaluate objectively. 

Previous research has primarily focused on disentangling style and content in text generation \cite{lee2021enhancing, zhu2023storytrans}, yet often encounters difficulties in preserving the original semantic content during the transformation process. More recent approaches have explored end-to-end models for text-style transfer \cite{dai2019style, lyu2023fine, roy2023conversation, tao2024cat}. Style Transformer \cite{dai2019style} does not impose prior assumptions on the latent representation of the source sentence, and instead employs an additional discriminator network to provide explicit style supervision. \citet{roy2023conversation} implements style transfer through style removal and few-shot in-context learning. CAT-LLM \cite{tao2024cat} introduces a pluggable Text Style Definition module that extracts stylistic features from input texts, enabling zero-shot style transfer through LLMs. This approach primarily considers lexical- and sentence-level attributes such as word length, part-of-speech distribution, syntactic structure, and emotion, among others.

\section{Method}

In this section, we introduce the Test-Time-Matching, which includes an automatic information extraction and analysis framework and a structured three-stage generation pipeline (shown in Figure \ref{fig:pipeline}). 

\subsection{Personality Analysis and Memory Mechanism}
\label{sec:s31}
We believe the characteristics that affect character dialogue response can be divided into two parts: cognitive tendency and linguistic style. Cognitive tendency determines what to say, while linguistic style determines how to say it. And cognitive tendency is largely shaped by the specific personalities and memories of the characters. In this work, we first extract the personalities and memories of specific characters from textual information by excluding the influence of linguistic style, in order to generate styleless responses to given input.

\textbf{Personality.} We begin by analyzing text chunks associated with the target character, employ the LLMs to generate brief descriptions of the character’s expressible personality traits for each chunk, and finally synthesize these into a comprehensive personality profile. We observed that LLMs tend to generate generic and risk-averse feature descriptions. As a result, imposing finer-grained constraints through carefully designed prompts becomes essential to preserve the diversity and specificity of the generated outputs. Without such guidance, the tendency toward neutrality can affect the quality and richness of the final responses.

\textbf{Memory.} As noted by \citet{llweng2023agent}, memory mechanisms, comprising both short-term and long-term components, are essential for enabling coherent and contextually grounded agent behavior. In this work, we regard the immediate chat history as the short-term memory, capturing the conversational context within a session. For long-term memory, we first create a structured representation of the character’s background by summarizing from textual chunks, including key biographical details such as age, gender, occupation, among others. Additionally, we construct an extended, graph-based vector database to encode and retrieve fine-grained personal traits, experiences, and relational information, thereby supporting a richer and more persistent form of long-term memory.

\begin{figure}[t]
\centering
\includegraphics[width=1\columnwidth]{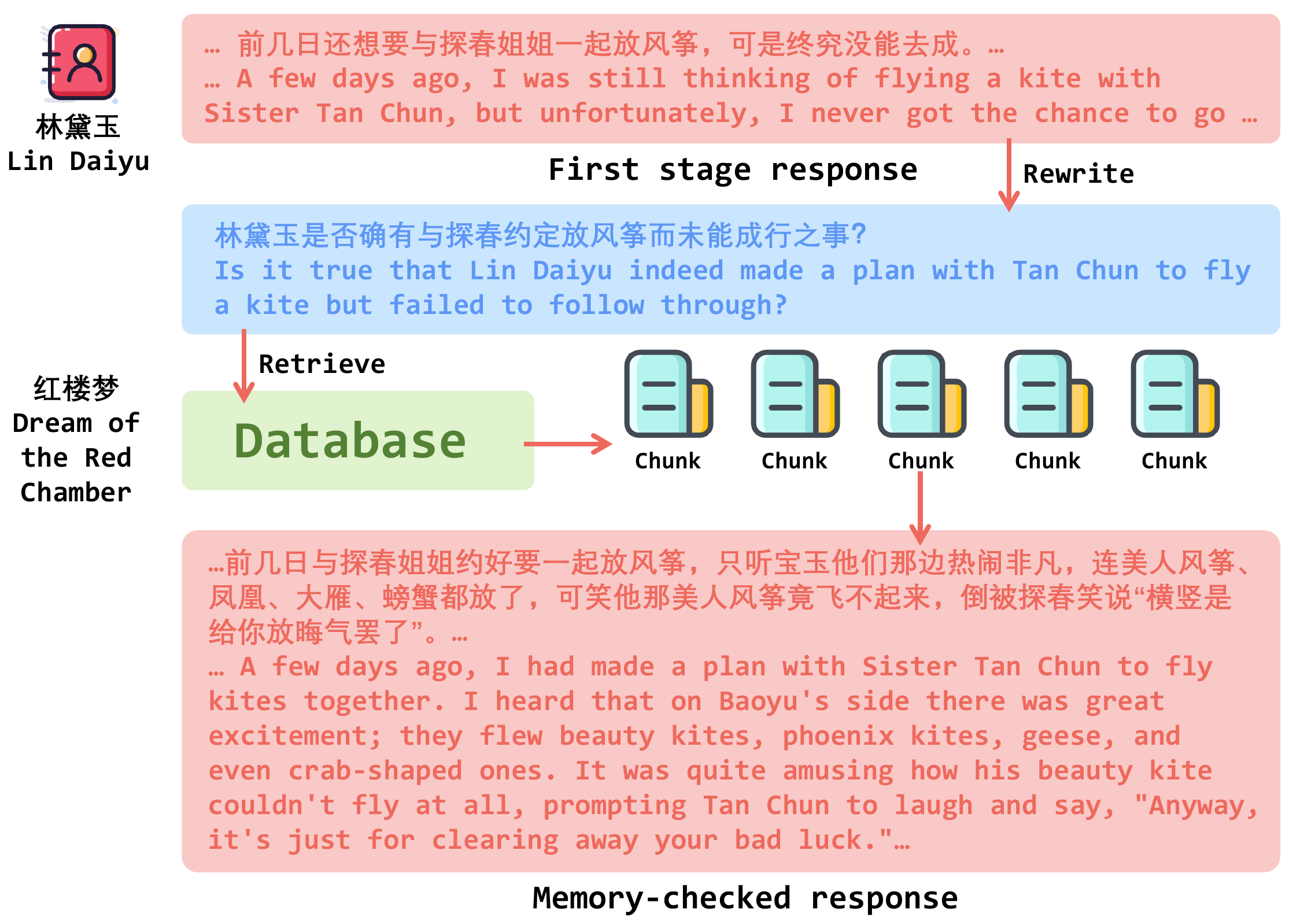} 
\caption{Memory-checked Response. In the second stage, TTM generates detailed query keywords by rewriting the first stage styleless response. These keywords are then used to retrieve relevant information from the graph-based database. The retrieved content is subsequently employed to correct or enrich the initial response, enhancing its accuracy and richness.}
\label{fig:memory_checking}
\end{figure}

\subsection{Styleless Response and Memory-checked Response Generation}
\label{sec:s32}

We aim to generate a styleless response to the user’s query based on the character’s personality and memory. Specifically, the personality traits and background information are provided as the system prompt to define the core characteristics of the LLM-based agent. To improve alignment, we also retrieve and incorporate the most relevant content from an external database into the input context, helping the model generate responses that better match the expected behavior of the character.

However, directly using the user's query as input makes it difficult to retrieve highly relevant content from the database. Therefore, before processing the query, we first generate a styleless response based only on the character’s personality and background. This initial response is then used to search the database for related knowledge. The related knowledge helps to supplement missing or incorrect details to get the second stage memory-checked response. Inspired by recent advances in query rewriting \cite{ma2023query}, a technique aimed at improving retrieval effectiveness by reformulating input queries, we guide LLM to identify potential knowledge errors in the first stage response. This process generates refined keywords that help to obtain high-relevant content from the database. This retrieved content is used to further correct and enhance the first-stage styleless response (shown in Figure \ref{fig:memory_checking}).

Although we have made every effort to minimize the influence of linguistic style in the character's personality and background descriptions, the content retrieved by RAG may still contain stylistically marked expressions. To address this, an additional linguistic style removal step can be applied to the output using an LLM, similar to \citet{tao2024cat} and \citet{roy2023conversation}, aiming to produce a cleaner, more neutral response. However, it should be noted that during the removal process, the LLM may change the original meaning of the sentence, particularly when dealing with complex or archaic language forms such as Classical Chinese, which are inherently more difficult to interpret accurately. Therefore, additional style removal needs to be used with caution.

\subsection{Linguistic Prompting and Progressive Matching}
\label{sec:s33}

We integrate the linguistic style of the character into the previously obtained memory-checked response by employing utterances retrieval combined with progressive matching, utilizing the capabilities of the LLM.

\textbf{Linguistic Prompting.} After generating the memory-checked response, the final step involves infusing the target character’s linguistic style to produce a stylized output. Drawing inspiration from CAT-LLM \cite{tao2024cat}, we represent a character’s linguistic style through two key components: linguistic preferences and common used words. Linguistic preferences include stylistic tendencies such as formal or informal tone, preference for long or short sentences, use of rhetorical questions, among others. Common used words are categorized and counted based on their part of speech. 

Additionally, since we already have the initial response, we can use it to retrieve semantically similar utterances from the character’s history conversations. These retrieved examples, which are semantically similar to the target input, serve as strong stylistic references to guide the LLM during the linguistic rewriting process. They help ensure the final output aligns closely with the character’s linguistic style.

\begin{figure}[t]
\centering
\includegraphics[width=1\columnwidth]{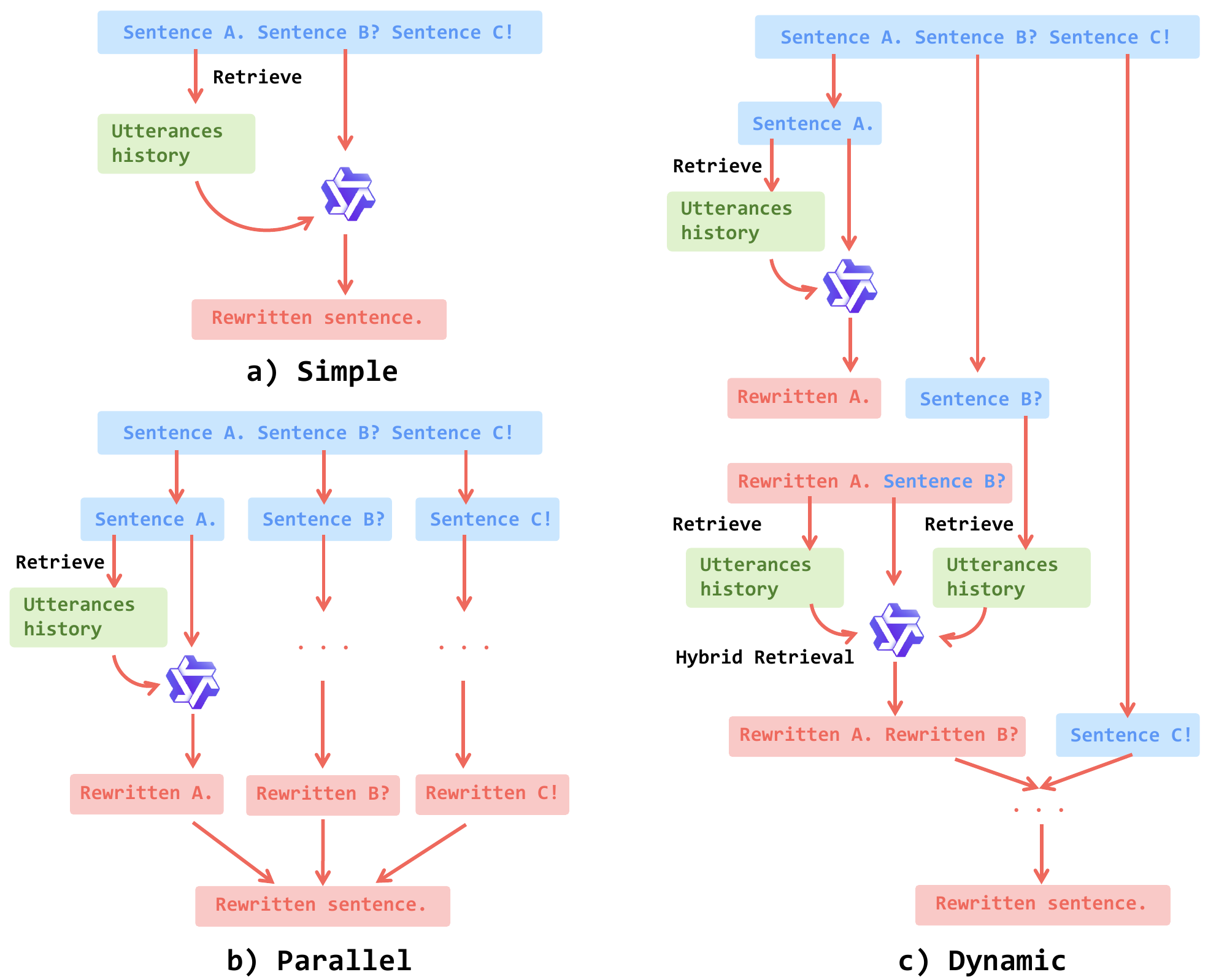} 
\caption{The details of progressive matching and hybrid retrieval. Compared to a) simple: simple rewriting, which directly reformulates queries in one step, we can use two enhanced strategies to improve the efficiency or quality. b) Parallel: parallel rewriting, where queries are split and rewritten to support parallel processing and speed up performance; c) Dynamic: dynamic rewriting, which adopts a progressive fusion approach to refine utterances iteratively and achieve higher similarity.}
\label{fig:prog_match}
\end{figure}

\textbf{Progressive Matching.} In the process of rewriting responses, we observed that the generated replies often include character actions and consist of multiple sentences, rather than being limited to single-phrase utterance. Such complex sentences are not ideal for query and rewriting tasks. They can make it harder to retrieve the most relevant segments. They may also lead to mismatches in style, which can reduce the overall quality of style transfer. To address this issue, we implement a segmentation-based approach in which each sentence is decomposed into individual components  (shown in Figure \ref{fig:prog_match}). These components are then rewritten sequentially from front to back through progressive matching. During each search step, the most recent non-stylized sentence added in the matching process is used as input to ensure contextual coherence and stylistic consistency. At the same time, character actions will not participate in the rewriting process and will be concatenated into the reply after the rewriting is completed.

\textbf{Hybrid Retrieval.} During each retrieval step of the progressive matching process, we can opt to combine the retrieval results from both the complete sentences that have been matched so far and the newly added sentence. This approach provides the LLM with a more coherent and contextually consistent reference template, facilitating improved stylistic alignment and overall rewriting quality.

During experiments, we found that the number of characters's history utterances available for linguistic style matching significantly impacts the results. When the number is insufficient, the LLM may overly consider stylistic adaptation in the presence of unrelated references, leading to unintended changes in the original meaning of the rewritten sentences.

\subsection{Implementation Details}
\label{sec:s34}
Our goal is to automatically construct RPLAs from unstructured textual sources. To achieve this, we begin by segmenting the input texts into manageable chunks. For each chunk, dialogues and their corresponding speakers are identified and extracted, resulting in a structured dataset that maps each speaker to their associated utterances. For each speaker, all text chunks are queried to retrieve those most relevant to the target speaker. Subsequently, the LLM-based agent is employed to analyze and characterize the speaker’s linguistic style.

Utilizing LLM agents, we progressively extract detailed personality traits and background information from the related chunks. In terms of background extraction, we focus on identifying salient and easily inferable attributes, including: name, gender, age, ethnicity, identity, occupation, physical appearance, health status, family background, historical context, key possessions, as well as interests and hobbies. This structured representation serves as the foundation for generating contextually and stylistically coherent character interactions.

\section{Experiments}

\begin{table*}
\small
\centering
% \scalebox{1}{
\begin{tabular}{lcccccccccc} 
\toprule[0.8mm]
Method & Base Model & \multicolumn{3}{|c}{LLM-as-Judge} & \multicolumn{3}{|c}{General Participants} & \multicolumn{3}{|c}{Linguistics Experts}\\
 & & \multicolumn{1}{|c}{C. P. $\uparrow$} & A. K. $\uparrow$ & Q. C. $\uparrow$ & \multicolumn{1}{|c}{C. P. $\uparrow$} & A. K. $\uparrow$ & Q. C. $\uparrow$ & \multicolumn{1}{|c}{C. P. $\uparrow$} & A. K. $\uparrow$ & Q. C. $\uparrow$ \\
\midrule[0.5mm]
\multicolumn{11}{l}{\textit{RPLA methods}}  \\
\midrule
ChatHaruhi & Qwen-3-32B & 8.6363 & 8.7272 & 8.6136 & 6.1250 & 6.4166 & 6.0416 & 5.8000 & 7.3500 & 5.8500 \\
% CharacterGLM & ChatGLM-6B & & & & & & & & & & & & \\
CoSER & LLaMA-3.1-70B & 7.4705 & 8.4705 & 7.5588 & 6.3055 & 6.6666 & 6.4444 & 6.0666 & 7.4666 & 5.8666 \\
\midrule
\multicolumn{11}{l}{\textit{Universal LLMs}}  \\
\midrule
GPT-4o & - & 7.7903 & 8.4838 & 7.8870 & 5.7916 & 6.1527 & 5.7916 & 4.9666 & 7.0666 & 5.2666 \\
% GPT-o3 & - & & & & & & & - & - & - & - & - & -  \\
Gemini-2.5-pro & - & 9.0967 & 9.5322 & 9.1451 & 6.9861 & 7.3750 & 6.9861 & 6.1000 & 7.4666 & \textbf{6.4000} \\
Qwen-3 & Qwen-3-235B-A22B & 8.6290 & 8.8870 & 8.6774 & 6.3750 & 6.5555 & 6.3611 & 5.9000 & 7.3666 & 6.3666 \\
Qwen-3 & Qwen-3-32B & 8.0322 & 8.4677 & 8.2096 & 6.5000 & 6.5694 & 6.4305 & 5.4333 & 7.3666 & 6.0000 \\
\midrule[0.5mm]
TTM (Ours) & Qwen-3-32B & \textbf{9.2096} & \textbf{9.5645} & \textbf{9.2741} & \textbf{7.2638} & \textbf{7.4861} & \textbf{7.0694} & \textbf{6.4333} & \textbf{7.6000} & 6.3000 \\
\bottomrule[0.8mm]
\end{tabular}
% \end{tabular}}
\caption{The evaluation results. For each dimension, the maximum score attainable is 10 and we highlight the best results in bold. Gemini-2.5-pro employs online grounding to enhance performance. Chain-of-thought (CoT) reasoning is not employed in all Qwen-based methods. The detailed evaluation records can be found in the supplementary materials.}
\label{evaluation_tab}
\end{table*}

In this section, we compare the role-playing performance of TTM against with other methods, including the specific RPLA methods (Prompting: ChatHaruhi \cite{li2023chatharuhi}, Training: CoSER \cite{wang2025coser}) and universal LLMs (GPT-4o \cite{hurst2024gpt4o}, Gemini-2.5-pro \cite{gemini25pro}, Qwen-3 \cite{yang2025qwen3}). We extract characters from classic Chinese and English novels and generate corresponding dialogues. 

\textbf{Metrics.}  We evaluate the dialogues based on three key criteria. \textbf{1) Consistency of Persona (C. P.)}, which assesses whether the conversational style and behavioral traits exhibited by the agent align with the played character. \textbf{2) Accuracy of Knowledge (A. K.)}, which evaluates whether the agent demonstrates accurate and contextually appropriate knowledge reflective of the character’s background, including both the correct articulation of relevant information and the avoidance of statements inconsistent with the character’s known attributes or setting. \textbf{3) Quality of Conversation (Q. C.)}, which captures raters’ overall impression of the conversation, encompassing aspects such as fluency, engagement, and their willingness to continue engaging in meaningful and in-depth dialogue with the agent. 

\textbf{Evaluation.} To evaluate the quality of dialogue generation across different methods, we conduct a comprehensive human assessment involving both \textbf{general participants} and \textbf{linguistics experts}, who are asked to rate multi-turn dialogues generated by various models for specific roles based on their subjective preferences. Additionally, we employ the \textbf{LLM-as-Judge} approach to provide automated evaluation for further analysis.

\textbf{Models.} TTM uses Graph-based RAG framework \cite{zhou2025depth} and LightRAG \cite{guo2024lightrag} to construct detailed memory database. The bge-large-zh-v1.5 \cite{bge_embedding} and the Qwen3-Embedding-0.6B \cite{qwen3embedding} are used for retrieval. During the experiments, we use Qwen-2.5-32B-Instruct \cite{qwen2025qwen25technicalreport} for information extraction and analysis. We select this model because Qwen-3, while more powerful, would extract more detailed information, resulting in higher memory and time requirements. By default, the dialogues are generated by Qwen-3-32B \cite{yang2025qwen3} without CoT reasoning.

\begin{figure}[t]
\centering
\includegraphics[width=1\columnwidth]{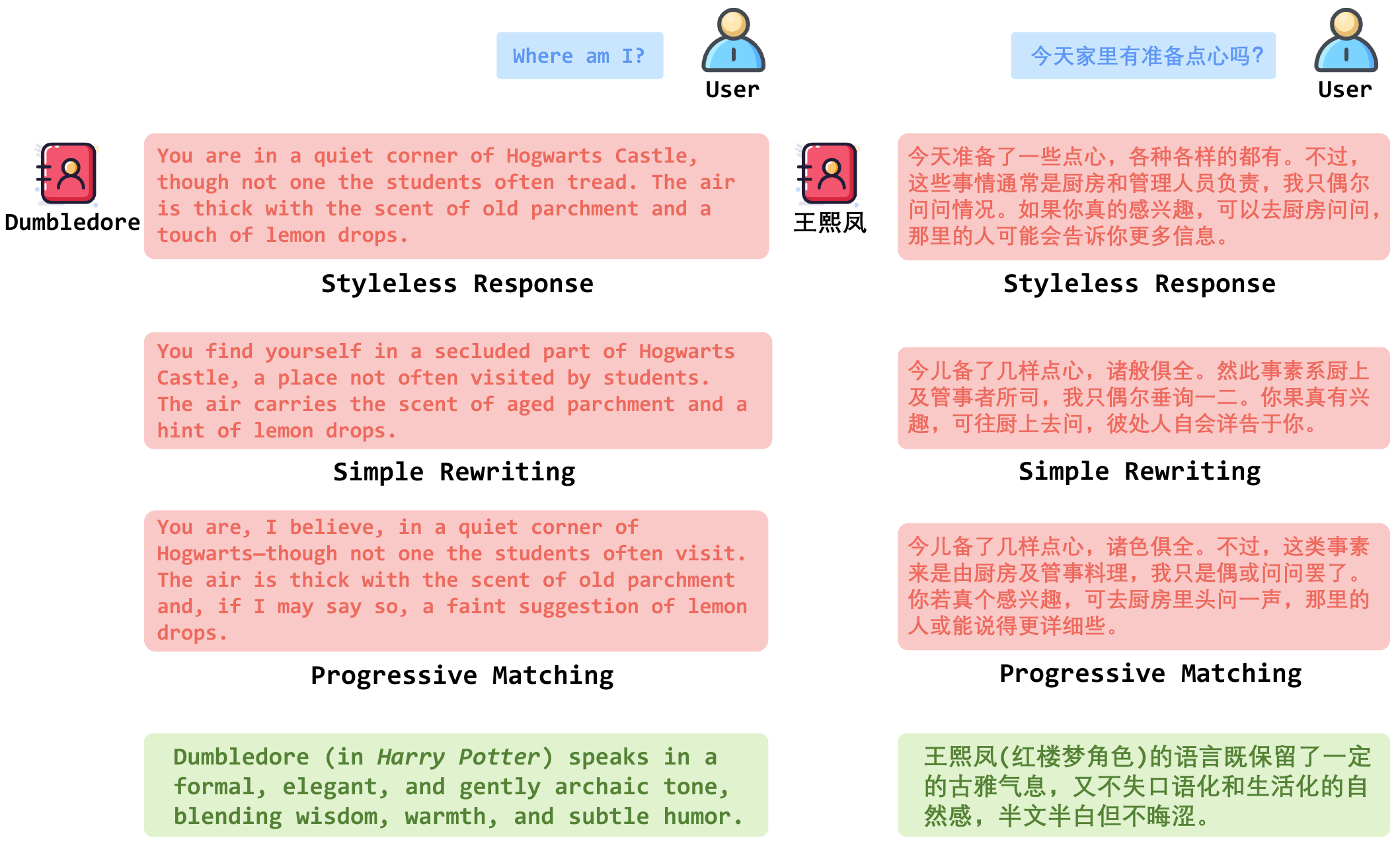} 
\caption{TTM achieves a high-fidelity reproduction of a target character’s linguistic style through progressive matching, thereby effectively enhancing the expressiveness and performance of RPLAs.}
\label{fig:prog_match_example}
\end{figure}

\subsection{Evaluation and Analysis}

Considering the character libraries of RPLA methods (ChatHaruhi and CoSER), we finally select 3 Chinese characters (Lin, Daiyu in \textit{Dream of the Red Chamber}, Duan, Yu and Xu, Zhu in \textit{The Demi-Gods and Semi-Devils}), and 3 English characters (Dumbledore and Hermione in \textit{Harry Potter}, and 
Elizabeth in \textit{Pride and Prejudice}) for evaluation. We conduct 4–6 conversational rounds for each character and method, with a single sample generated per character and method. Due to training data limitations, CoSER only generates dialogues for English characters. As a result, each questionnaire contained 37 dialogue samples. For each sample, raters are required to rate the content along three dimensions: Consistency of Persona (C. P.), Accuracy of Knowledge (A. K.), and Quality of Conversation (Q. C.). Ratings are assigned on a scale from 0 to 10, with 10 being the highest score.

\begin{figure*}[t]
    \centering
    \begin{subfigure}[b]{0.33\textwidth}
        \centering
        \includegraphics[width=\textwidth]{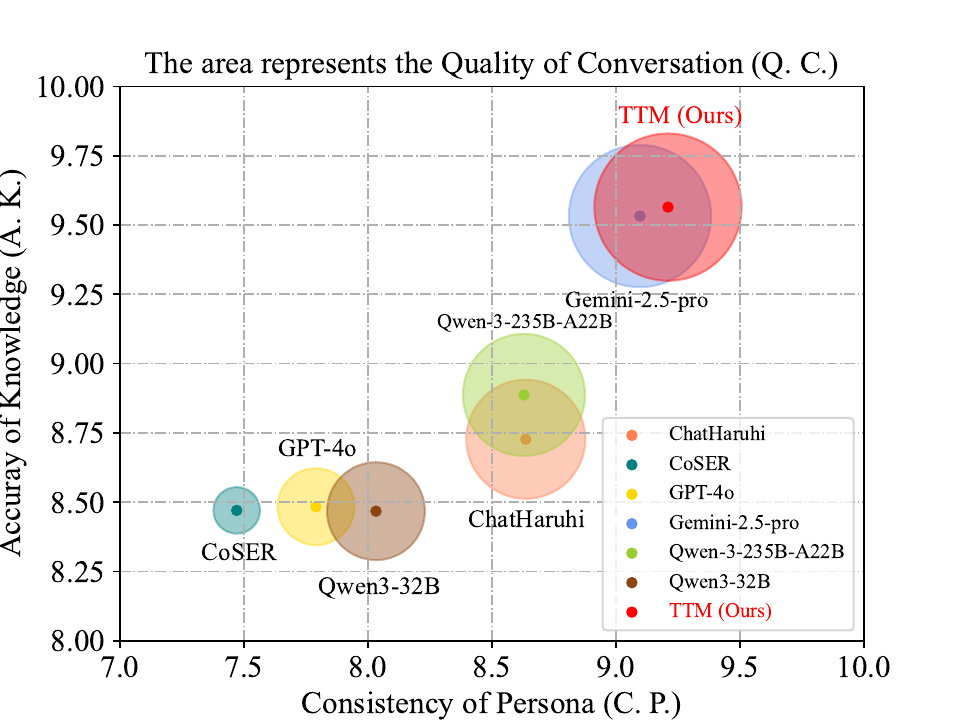}
        \caption{LLM-as-Judge.}
        \label{fig:sub1}
    \end{subfigure}
    \hfill
    \begin{subfigure}[b]{0.33\textwidth}
        \centering
        \includegraphics[width=\textwidth]{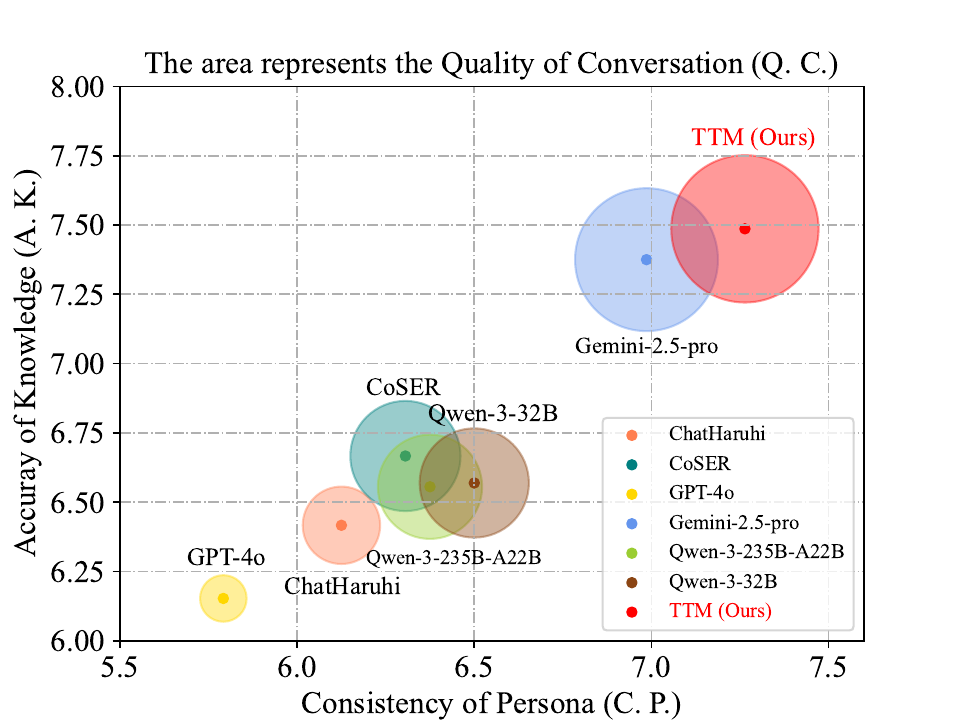}
        \caption{General Participants.}
        \label{fig:sub2}
    \end{subfigure}
    \hfill
    \begin{subfigure}[b]{0.33\textwidth}
        \centering
        \includegraphics[width=\textwidth]{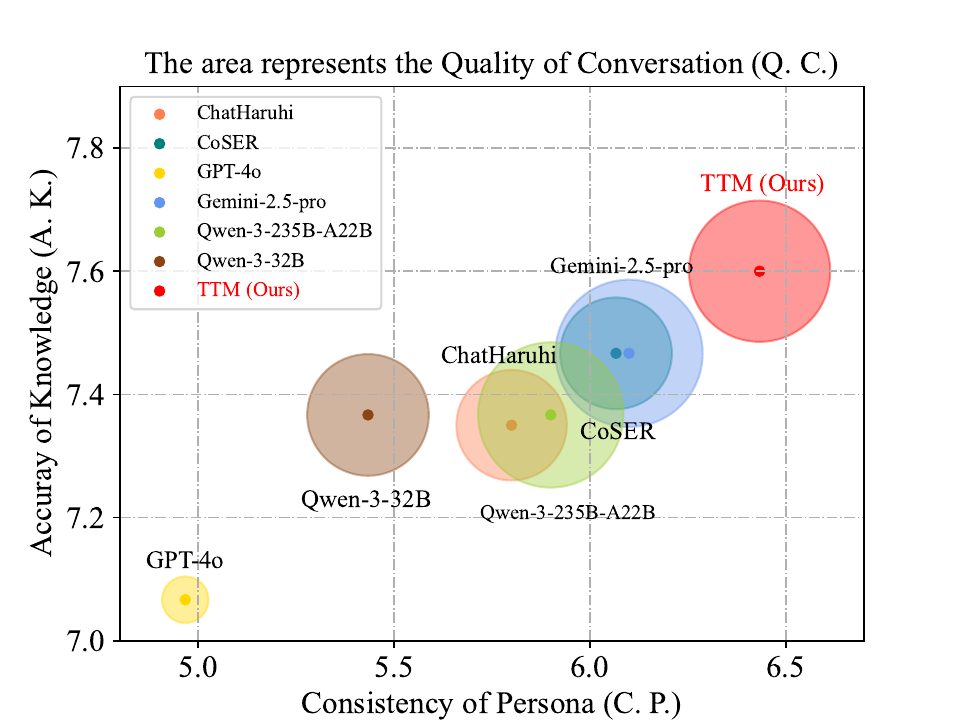}
        \caption{Linguistics Experts.}
        \label{fig:sub3}
    \end{subfigure}
    \caption{Visualization of the evaluation results.}
    \label{fig:group_visualization}
\end{figure*}

\textbf{Human Assessment.}
We collect 12 responses from the general participants and 5 responses from the linguistics experts (both Chinese and English). On average, each rater needs to spend more than 50 minutes to complete the evaluation. The evaluation results are shown in Table \ref{evaluation_tab}. We also provide the corresponding visualization results in Figure \ref{fig:group_visualization}.

The results indicate that TTM demonstrates strong overall performance, as reflected in the evaluations from both general participants and linguistic experts. It is worth noting that the baseline model Qwen-3-32B used in our study showed relatively poor performance in the evaluations, which further highlights the effectiveness of TTM. By leveraging test-time scaling and context engineering, TTM significantly improves the model's performance in role-playing tasks without fine-tuning. 

However, when it comes to conversation quality, TTM ranks slightly below Gemini-2.5-pro and Qwen-3-235B-A22B in assessment conducted by linguistic experts. From raters' feedback, we learned that TTM's long responses caused reading fatigue. Although our responses contained rich role-related emotions and knowledge, they felt unnatural in everyday conversation, which hurt the overall experience. This happened because, during the questionnaire sampling phase, we did not limit the response length to allow for richer information. As a result, during the memory-checking stage, many details were added based on retrieved knowledge. In practice, this can be improved by adding a simple summarization module after the memory-checking stage, which would also greatly reduce the resource and time costs of progressive matching.

\textbf{LLM-as-Judge.} For LLM-as-Judge, we employ GPT-4.1 \cite{gpt-4.1} with the temperature of 0.2 and top-p of 0.8. The prompt is adapted from MMRole \cite{dai2024mmrole} with minor changes. For each comparison, we input two examples to GPT to obtain a pair of scores. To enhance evaluation robustness, we also swap the input positions to generate a mirrored result. The results are shown in Table \ref{evaluation_tab} and Figure \ref{fig:group_visualization}.

The overall results align with human preferences. However, LLMs tend to give higher scores compared to human raters when evaluated in absolute terms. Here also shows a significant difference between LLM-as-judge and human ratings for CoSER. This may be due to two factors. First, CoSER's responses are much shorter than those of other methods. Humans may prefer shorter interactions, while LLMs tend to favor longer, more informative dialogues. Second, shorter responses include less role-related knowledge. LLMs, due to their tendency to generate hallucinations, may not detect incorrect knowledge well and instead focus on the richness of the content. In contrast, human raters may notice more errors or inconsistencies in longer responses, which can lower their overall rating. These two factors value careful attention in future research on LLM-as-Judge methodologies.

\textbf{Linguistic Style.} Inspired by adversarial detection, we also conduct a user study to verify the linguistic style improvement of TTM. For each question, participants are asked to select the sentence generated by LLMs, from among three examples drawn from the role’s historical utterances in books. The historical utterances are selected by performing vector similarity matching between the inserted LLM-generated sentences and the role's historical database. In order to reduce the difficulty of participants' selection, we use Qwen-2.5-32B-Instruct \cite{qwen2025qwen25technicalreport} with weaker performance as the base model. We collected a total of 21 responses. TTM was identified with a probability of $15.48\%\:(78/504)$, while Qwen-2.5-32B-Instruct was identified with a probability of $24.80\%\:(125/504)$. This indicates that, for humans, identifying the linguistic style of a specific role remains a challenging task. However, the significantly lower selection rate of TTM demonstrates that our method effectively enhances consistency with the target role's linguistic style (shown in Figure \ref{fig:prog_match_example}).

\subsection{Limitations}
Although TTM achieves notable advantages, it still faces several clear limitations. The three-stage generation will lead to additional computation costs in test-time. But at the same time, it can also enable small parameter models to have role-playing capabilities comparable to or even better than larger scale models. The automatically extracted information may contain descriptions that do not match the characters, but since the profiles are explicitly stored in plain text, it can be easily deleted or modified. Additionally, progressive matching of sentences sentence-by-sentence may occasionally compromise inter-sentential coherence. More discussions are conducted in the Appendix.

\section{Conclusion}

In this work, we propose a fully automated framework for generating RPLAs from textual inputs, which incorporates a structured, three-stage dialogue generation pipeline. Our approach leverages test-time scaling and context engineering to enhance both the controllability and expressive richness of role-playing interactions. The proposed method can be applied to create highly realistic role-playing experiences or to generate high-quality dialogue datasets. Despite the advancements demonstrated, the deployment of RPLAs raises significant ethical concerns that warrant careful consideration. These include challenges related to identity representation, the potential influence on user behavior, and the risk of generating harmful or inappropriate content. While broader societal implications remain an important area for further discussion, the integration of test-time scaling and context engineering presented in this study offers valuable insights that could inspire innovation across other domains of LLM applications, thereby releasing their full potential.

\section{Acknowledgments}
We sincerely thank the participants who generously contributed their time to our user study. We also gratefully acknowledge Xiang Li for sharing the visualization code. This work is supported by the National Natural Science Foundation of China (62032011) and the Natural Science Foundation of Jiangsu Province (BK20211147).

\bibliography{aaai2026}

% \clearpage
% \setcounter{page}{1}

% \setcounter{section}{0}
% \setcounter{figure}{0}
% \setcounter{table}{0}

% \renewcommand\thesection{\Alph{section}}
% \renewcommand\thetable{S\arabic{table}}
% \renewcommand\thefigure{S\arabic{figure}}

% \appendix
% \section*{Appendix}
\section*{\fontsize{15}{40}\selectfont Appendix}

\section{More Details}

\subsection{Extraction and Analysis}

\textbf{Background.} We extract the required background information from each chunk. As repeated information may be extracted from different chunks, we perform a summarization process every five extraction steps. This summarization is conducted separately for each individual attribute in the background.

\textbf{Memory Database.} We construct the memory database using the entire book, rather than relying solely on role-related information. This approach is adopted to avoid the high computational and storage costs associated with building separate databases for each individual role.

\textbf{Historical Utterances.} Since roles and dialogues extraction are performed by the LLM on a per-chunk basis, the model may assign different names to the same character due to lack of complete context. To address this, we employ a post-processing step to merge entities referring to the same character. However, this process may still lead to potential omissions. Therefore, at runtime, users can include historical utterances involving multiple entities that refer to the same character.

\subsection{Implementation}

\textbf{Chunk Size.} The size of the split chunks directly affects both the quality of the final response and the processing time. Larger chunks may capture more context but increase computational cost, while smaller chunks may lead to fragmented understanding. By default, we set the chunk size to 512 tokens with an overlap of 64 tokens to balance context coverage and efficiency.

\textbf{Utterances Retrieval.} We employ a hybrid retrieval approach combining BM25 \cite{robertson1994some} and the Qwen-3-Embedding-0.6B model \cite{qwen3embedding} to identify historical utterances similar to the given query. The retrieval weights for the two methods are set to [0.5, 0.5], ensuring a balanced contribution from both lexical and semantic matching.

\textbf{Prompts.} To achieve consistent performance across languages, we recommend using prompts that match the input language.

\subsection{Customization}

All modules in the implementation of TTM are independently decoupled. This design allows for flexible customization within the complete TTM pipeline. 

For instance, the entire Memory-checking module can be removed to avoid the computational and storage overhead associated with building and maintaining the database. Alternatively, the Style Generation module can be moved before the Memory-checking module, preventing the introduction of excessive details that could significantly increase the computational cost of progressive matching.

The explicitly stored character features (personality, background, linguistic style, and historical utterances) can also be manually modified as needed to create a role that truly aligns with the user's specific requirements.

\section{More Discussions}
Though TTM reaches the best performance when againsting with other methods. We observe some phenomenons as well. Here we are providing a more detailed discussion of our experiments.

\subsection{The Influences of Base Model} 
As a test-time method, TTM is fundamentally dependent on the capabilities of the base model. In our experiments, we observed a notable performance gap between Qwen-3-32B \cite{yang2025qwen3} and Qwen-2.5-32B-Instruct \cite{qwen2025qwen25technicalreport}, with the former significantly outperforming the latter. It remains unclear whether TTM utilizes the system prompt, which may have influenced the performance of Qwen-2.5-32B-Instruct.

To investigate this, we evaluate the role-playing capabilities of Qwen-2.5-32B-Instruct using both system prompts and standard prompts for comparison. The results indicate that the model's performance is considerably better when role-playing is induced through standard prompts alone, suggesting the potential risks in how the system prompt is processed or leveraged for different models.

Another influential factor related to the base model is its inherent knowledge acquired during pretraining. Although TTM is a training-free method, if the base model has previously encountered data related to the target character during its initial training, it can significantly enhance the quality of the model's responses during the styleless phase of TTM. This, in turn, leads to an overall improvement in the final performance.

\subsection{Retrieval Quality} 
In the TTM process, retrieval is employed at two key stages. The first involves querying relevant chunks from the memory database using the rewritten query. The second entails retrieving similar sentences from the character's historical dialogue database based on the generated response. The quality of retrieval at both stages significantly impacts the overall performance of TTM.

\subsection{Information Overload} 
In some responses, it is not necessary to retrieve additional information from memory to enhance consistency or expressiveness. Applying memory-checking to every utterance may lead to an overload of memory-related content or the incorporation of irrelevant knowledge. This can result in contextual incoherence. Moreover, it may lead to the repetitive use of multiple paragraphs to convey the same piece of information in the final response. All of these factors can notably affect the user's conversational experience.

\subsection{Knowledge Misuse} 
To reduce computational and storage overhead, we construct a single memory database per book rather than maintaining separate databases for individual characters. This approach may lead to potential knowledge misuse issues, including commonsense errors (e.g., incorrect forms of address or relationships) and memory attribution errors, such as incorporating narrative or third-person perspective information that does not belong to the character.

\subsection{Timeline of Memory} 
A critical issue affecting dialogue experience is the temporal disorganization of the memory database used for retrieval, whereas role-related information in books is inherently linear and ordered. Existing RAG methods consider only relevance to the query, ignoring temporal sequence. Consequently, during dialogue, a role may reference later memories first and then abruptly revert to earlier ones, leading to a disjointed experience for users familiar with the role's background.

\subsection{Time and Computational Resource Consumption} 
For each book, the primary computational costs are attributed to dialogues extraction and database construction. We run Qwen-3-32B on four NVIDIA 3090 GPUs; processing a book of approximately 2MB in size takes 2-3 hours.

In addition, extracting the role's background information takes between several tens of minutes to approximately 3 hours, depending on the number of role-related chunks.

During conversation, styleless response, query rewriting, memory-checked response, and short response summarization (if needed) each require one LLM call. Memory-checked response also involves one retrieval step. The number of LLM calls for linguistic rewriting depends on the selected mode and the length of input.

\section{More Results}

The complete response generation process in TTM (based on Qwen-3-32B) are presented here. Considering both the space considerations and the feedback of expert raters, we add a summarization step following the memory-checking module during generation to avoid producing excessively long responses. \\ \\

\subsection{Dumbledore in Harry Potter}
\textbf{User:} 

Ugh, my head hurts. Where am I? \\

\noindent
\textbf{Styleless Response:} 

(You find yourself sitting up slowly, the soft crackle of a nearby fire the only sound in the quiet room. A warm, familiar voice speaks gently.) You are safe, my dear. In the Headmaster’s office at Hogwarts. 

(A long white beard ripples slightly as the speaker nods.) Let me fetch you some tea and a restorative draught. You’ve had a nasty knock to the head, but no lasting harm—none that cannot be mended with care and time. \\

\noindent
\textbf{Memory-checked Response:} 

(You sit up slowly, the fire crackling softly nearby. A warm, familiar voice speaks gently.) You are safe, my dear—here in the Headmaster’s office at Hogwarts.

(A long white beard ripples slightly as the speaker nods.) Let me bring you some tea and a restorative draught. Minerva found you near the Whomping Willow, and you’ve been unconscious for hours.

(The eyes twinkle, soft with concern.) Dark magic stirs again, and the world beyond these walls grows uncertain. But here, you are safe—for now.

(A gentle hand rests on your shoulder.) Rest. I will not let harm come to you. Not while I still draw breath. \\

\noindent
\textbf{Stylized Response:}

(You sit up slowly, the fire crackling softly nearby. A warm, familiar voice speaks gently.) You are quite safe, I assure you—here in my office at Hogwarts.  

(A long white beard ripples slightly as the speaker nods.) Let me fetch you some tea and a restorative draught, my dear. Minerva found you near the Whomping Willow, and you have been unconscious for some hours.  

(The eyes twinkle, soft with concern.) Dark forces stir once more, and the world beyond these walls grows ever more uncertain. But here, you are safe—for now.  

(A gentle hand rests on your shoulder.) Rest now, my dear. I shall see to it that no harm comes to you — not while I still have breath in my body.

-------------------------------------------------------------------

\noindent
\textbf{User:}

Oh thank you, sir. But who are you? \\

\noindent
\textbf{Styleless Response \& Memory-checked Response:} 

(Adjusting half-moon spectacles with a faint smile) Albus Dumbledore, Headmaster of Hogwarts. You may call me Professor Dumbledore. I have been expecting you, in a way — though not under such troubling circumstances. 

(The eyes gleam with quiet warmth.) Rest easy, and I shall explain all in time. \\

\noindent
\textbf{Stylized Response:}

(Adjusting half-moon spectacles with a faint smile) Albus Dumbledore, Headmaster of Hogwarts. You will find it quite proper to call me Professor Dumbledore. In a manner of speaking, I have been expecting your arrival — though not, I must say, under such troubling circumstances.  

(The eyes gleam with quiet warmth.) Fear not, my dear boy — all shall be made clear in good time. \\

-------------------------------------------------------------------

\noindent
\textbf{User:}

Ah Professer Dumbledore, thanks for saving me. But what happend? \\

\noindent
\textbf{Styleless Response:} 

(Smiling gently, though with a gravity that underlines the mirth in his eyes) "Ah, my dear boy, it is I who must thank you—for your courage, for your presence here. As for what has happened... let us say, you have stumbled into a most curious tangle of events. Some of which, I confess, even I am still unraveling.  

(He gestures toward a nearby chair, his long fingers tapping thoughtfully on the armrest.) But first, let us speak of what you remember. And perhaps we shall find, together, the truth that lies beneath." \\

\noindent
\textbf{Memory-checked Response:} 

(Sitting with a thoughtful air, his silver beard slightly tousled, Dumbledore’s eyes glimmer with a quiet intensity.) "Ah, my dear boy, it is I who must thank you—for your courage, for your presence here. As for what has happened... let us say, you have stumbled into a most curious tangle of events.  

(He gestures toward a nearby chair, his long fingers tapping thoughtfully on the armrest.) You have seen things that are not pleasant. And there is no countercurse. It seems you ran over the details in your mind, trying to make sure you could remember them. That is wise. Memory is our greatest tool in such times. But first, let us speak of what *you* remember. And perhaps we shall find, together, the truth that lies beneath." \\

\noindent
\textbf{Stylized Response:}

(Sitting with a thoughtful air, his silver beard slightly tousled, eyes glimmering with quiet intensity.) Ah, my dear boy, it is I who should be thanking *you*—for the courage you have shown and for being here at all. As for what has happened—let us say, you have found yourself entangled in a most curious and perilous set of circumstances.  

(Gesturing toward a nearby chair, long fingers tapping thoughtfully on the armrest.) I fear you have witnessed things of a most distressing nature, and I must tell you, there is no countercurse. I suspect you have most carefully reviewed the events, seeking to fix them in your mind. That is wise indeed. Memory, my dear boy, is our greatest tool in such trying times. But first, let us speak of what *you* remember. And perhaps, together, we shall uncover the truth that lies beneath. \\

-------------------------------------------------------------------

\noindent
\textbf{User:}

I can't remember anything, I just remember there was a snake. \\

\noindent
\textbf{Styleless Response:} 

(Leaning forward slightly, eyes alight with quiet concern and curiosity.) A snake, you say? 

(The hand reaches out, fingers hovering gently over the air as if tracing an unseen thread.) How peculiar. A snake, indeed—though I fear it may not have been what it seemed. You are not the first to see it, nor shall you be the last.  

(Adjusting the half-moon spectacles, the voice softens.) You were not alone in that moment. Others have walked this path before you, and others will follow. But you—

(A pause, and the eyes twinkle, though now with a deeper, more solemn light)—you have already done more than most could ever hope to do. Let us begin at the beginning, then. Tell me what you can, however faint the memory. Even the smallest thread may lead us to the truth. And I, for one, have no intention of letting this mystery remain unsolved. \\

\noindent
\textbf{Memory-checked Response:} 

(Leaning forward slightly, eyes alight with quiet concern.) A snake, you say? 

(The hand reaches out, fingers hovering gently.) How peculiar. A snake—though I fear it may not have been what it seemed. You were not alone in that moment. Others have walked this path before. But you—

(A pause, eyes twinkling with a deeper light)—you have already done more than most. Let us begin at the beginning. Tell me what you can. Even the smallest thread may lead us to the truth. The Chamber of Secrets was opened once before, and the consequences were dire. The serpent within was silenced—at least for now. As for the Horcrux... such things are not easily undone. They linger like a shadow. The artifact must be found and dealt with before it brings further harm. So, tell me—what do you remember of the snake? Was it large, or small? Did it speak, or simply watch? \\

\noindent
\textbf{Stylized Response:}

(Leaning forward slightly, eyes alight with quiet concern.) Ah, a snake, I see?  

(The hand reaches out, fingers hovering gently.) How very curious, indeed. A snake, you say? I must confess, I have long suspected that such creatures are not always what they seem. You were not the first to walk that path—others have found themselves upon it before you. But you—  

(A pause, eyes twinkling with a deeper light) I am afraid, my dear, that you have carried a burden most would not survive, even in their boldest hour. If you are prepared, let us begin at the beginning. Tell me all that remains with you—no matter how faint or wrapped in darkness—for even the smallest recollection may prove of the utmost importance. The Chamber of Secrets was opened once before, and the grief it brought was great. The serpent that lies within slumbers—for now. And as for the Horcrux… such things do not vanish so easily. They remain, like a shadow that clings when the light is gone. The object must be found and faced, or else it shall bring sorrow once more. Tell me, then—what do you recall of the serpent? Was it of great size, or small? Did it speak, or merely watch? \\

-------------------------------------------------------------------

\noindent
\textbf{User:}

I... I’m not sure. It... he had red eyes, and he could speak. Could I—could I maybe have something to drink, professor? \\

\noindent
\textbf{Styleless Response:} 

(Nodding gravely, though his voice remains gentle as the rustle of parchment.) Of course, my dear boy. Of course.  

(A wand flicks, and a steaming cup of tea appears beside you, accompanied by a small vial of shimmering blue liquid.) Alcohol, I’m afraid, is a poor substitute for the restorative draught I have prepared for you. 

(He settles back into his chair, fingers steepled, eyes never leaving yours.) The serpent spoke, you say. Red eyes—yes, I have heard of such eyes before. A mark of great power, and great danger. But you are safe now. That is what matters.  

(He sips his own tea, the firelight catching in his beard.) Tell me, what else lingers in your memory? Even the smallest detail may guide us forward. \\

\noindent
\textbf{Memory-checked Response:} 

(Leaning forward slightly, his eyes steady and kind.) Of course, my dear boy—of course. 

(A flick of the wand, and a cup of tea and a small vial appear.) A warm cup of tea, perhaps with elderflower or chamomile, would be better suited to your condition. If you must have something stronger, a glass of mead — milder than firewhisky.

(A pause.) Yes, I have heard of such eyes. A mark of power—and danger. But you are safe now. Tell me, what else do you remember? Even small details may help. Hogwarts is old, and it remembers. If something has stirred, we must know why. \\

\noindent
\textbf{Stylized Response:}

(Leaning forward slightly, eyes steady and kind.) Of course, my dear boy — I shall fetch you something at once.  

(A flick of the wand, and a cup of tea and a small vial appear.) A cup of tea, perhaps with a touch of elderflower or chamomile, might be a most suitable choice at this hour. Should you still feel the need for something stronger, I would suggest a glass of mead — a drink of milder temperament, I should imagine, than firewhisky.

(A pause.) I am quite certain I know those eyes — a most disquieting omen, I fear, of great strength and danger. But you are safe now, I assure you. Tell me, child, what else do you recall? Even the smallest memory may yet prove of the utmost importance. Hogwarts is ancient, and it remembers. If something has stirred within these walls, we must know why. \\

-------------------------------------------------------------------

\noindent
\textbf{User:}

I’m sorry, sir… I can’t remember anything else right now. If I could just rest a bit, maybe I’ll remember more tomorrow. \\

\noindent
\textbf{Styleless Response:} 

(Sitting back, the twinkle in the eyes softening like a lullaby.) Of course, my dear boy. Of course. Rest is the wisest course when the mind is weary. You have faced more than most your age could bear. Take all the time you need.  

(A gentle wave of the wand, and a warm, silvery light washes over the room, dimming the fire’s glow.) I shall be here when you awaken. And if you dream—should the serpent return in your sleep—do not fear. I shall be watching. Always.  

(Softly, with a voice like the rustling of old parchment.) Sleep now. The world will wait. And so shall I. \\

\noindent
\textbf{Memory-checked Response:} 

(Sitting back, the twinkle in the eyes softening.) Of course, my dear boy. Rest is the wisest course when the mind is weary. You have faced more than most your age could bear. Take all the time you need.  

(A gentle wave of the wand dims the fire’s glow with a warm, silvery light.) I shall be here when you awaken. And if you dream—should the serpent return in your sleep—do not fear. I shall be watching. Always. The serpent, I daresay, is not merely a dream. It is a shadow from the past—perhaps even a whisper from the Chamber of Secrets. You are connected to it in ways you have yet to understand.  

(Softly, like the rustling of old parchment.) Sleep now. The world will wait. And so shall I. \\

\noindent
\textbf{Stylized Response:}

(Sitting back, the twinkle in the eyes softening.) Of course, my dear boy — rest well, for I would say that rest is indeed the wisest course when the mind is weary. May clarity return to you with the morning light. You have faced more than most of your age could bear. Take all the time you need.  

(A gentle wave of the wand dims the fire’s glow with a warm, silvery light.) I shall remain by your side until you awaken. Should the serpent find its way into your dreams, know that I shall be watching, always. I would suggest that the serpent is more than mere fancy. It is a shadow of the past — perhaps even a whisper from the Chamber of Secrets. You are connected to it in ways you have yet to understand. 

(Softly, like the rustling of old parchment.) Rest now, my dear boy. Sleep may bring clarity when the time is right. The world will wait — and so shall I. \\

\end{document}